\documentclass[runningheads,a4paper]{llncs}

\usepackage{amsmath}
\usepackage{amssymb}
\usepackage[misc]{ifsym}
\usepackage{graphicx}
\usepackage{color}
\usepackage{algorithm}
\usepackage{algorithmicx}
\usepackage{algpseudocode}
\usepackage{subfigure}
\usepackage{multirow}
\usepackage{array}
\newcolumntype{L}[1]{>{\raggedright\let\newline\\\arraybackslash\hspace{0pt}}m{#1}}
\newcolumntype{C}[1]{>{\centering\let\newline\\\arraybackslash\hspace{0pt}}m{#1}}
\newcolumntype{R}[1]{>{\raggedleft\let\newline\\\arraybackslash\hspace{0pt}}m{#1}}

\usepackage{url}
\urldef{\mailsa}\path|{alfred.hofmann, ursula.barth, ingrid.haas, frank.holzwarth,|
\urldef{\mailsb}\path|anna.kramer, leonie.kunz, christine.reiss, nicole.sator,|
\urldef{\mailsc}\path|erika.siebert-cole, peter.strasser, lncs}@springer.com|

\begin{document}

\mainmatter  % start of an individual contribution

% first the title is needed
\title{Infinite Curriculum Learning for Efficiently Detecting Gastric Ulcers in WCE Images}

% a short form should be given in case it is too long for the running head
\titlerunning{Infinite Curriculum Learning for Efficiently Detecting Gastric Ulcers}

% the name(s) of the author(s) follow(s) next
%
% NB: Chinese authors should write their first names(s) in front of
% their surnames. This ensures that the names appear correctly in
% the running heads and the author index.
%
\author{Xiaolu Zhang\thanks{This work was done when Xiaolu Zhang worked at IBM Research - China.}\inst{1} \and
Shiwan Zhao\inst{2} \and
Lingxi Xie\inst{3}}
\authorrunning{ZHANG et al.}
% (feature abused for this document to repeat the title also on left hand pages)

% the affiliations are given next; don't give your e-mail address
% unless you accept that it will be published

\institute{Ant Financial, Hangzhou, China \\
\email{yueyin.zxl@antfin.com} \and
IBM Research - China \\ \email{zhaosw@cn.ibm.com} \and
The Johns Hopkins University, Baltimore, USA\\
\email{198808xc@gmail.com}}

%
% NB: a more complex sample for affiliations and the mapping to the
% corresponding authors can be found in the file "llncs.dem"
% (search for the string "\mainmatter" where a contribution starts).
% "llncs.dem" accompanies the document class "llncs.cls".
%

\toctitle{Infinite Curriculum Learning for Efficiently Detecting Gastric Ulcers}
\tocauthor{Anonymous Authors}
\maketitle

\begin{abstract}
The Wireless Capsule Endoscopy (WCE) is becoming a popular way of screening gastrointestinal system diseases and cancer. However, the time-consuming process in inspecting WCE data limits its applications and increases the cost of examinations. This paper considers WCE-based gastric ulcer detection, in which the major challenge is to detect the lesions in a local region. We propose an approach named {\bf infinite curriculum learning}, which generalizes curriculum learning to an infinite sampling space by approximately measuring the difficulty of each patch by its {\em scale}. This allows us to adapt our model from local patches to global images gradually, leading to a consistent accuracy gain. Experiments are performed on a large dataset with more than $3$ million WCE images. Our approach achieves a binary classification accuracy of $87\%$, and is able to detect some lesions mis-annotated by the physicians. In a real-world application, our approach can reduce the workload of a physician by $90\%$--$98\%$ in gastric ulcer screening.
\end{abstract}

\section{Introduction}
\label{Introduction}

Gastrointestinal (GI) system cancer has become a major threat to human lives with the improvement of diet~\cite{Siegel2017}. The early diagnosis and treatment of GI diseases are fundamental in reducing the death rate of GI system cancer. However, many GI diseases do dot have specified symptoms~\cite{Li2010china}. Although conventional gastroscopy check provides a standard of diagnosis, the high requirement of equipments and the low acceptance degree in population (mainly caused by the pain brought to the patients by this check) limit its application in screening GI diseases. In comparison, the Wireless Capsule Endoscopy (WCE) offers a low-risk, non-invasive visual inspection of the patient's digestive tract compared to the traditional endoscopy~\cite{Karargyris2011}. It has been recommended as a first-line examination tool to be used during a routine examination~\cite{Liao2016}.

One of the main challenges in the WCE-based examinations is the sparsity of useful information. For each patient, around $55\rm{,}000$ pictures are taken in the patient's digestive track, but the evidences of abnormalities ({\em e.g.}, Crohn’s disease, ulcers, blood-based abnormalities and polyps, {\em etc.}) only appear in a few of them. It takes a physician $45$ minutes to several hours to finish the visual analysis~\cite{Karargyris2009}, yet the performance is far from satisfaction~\cite{Zheng2012}. This raises the importance in designing an AI system to reduce the burden of the physicians,  the cost of the WCE-based examinations, and improve the accuracy. The topic of automatically inspecting WCE data has attracted a lot of attentions in the conventional image processing  area~\cite{Karargyris2011}\cite{Li2009}\cite{SuZhang2009}\cite{Yeh2014}. Recently, the fast development of deep learning especially convolutional neural networks~\cite{krizhevsky2012} has brought us a new opportunity in boosting the performance of analyzing WCE data.

\newcommand{\figurewidth}{12cm}
\begin{figure}[t]
\begin{center}
    \includegraphics[width=\figurewidth]{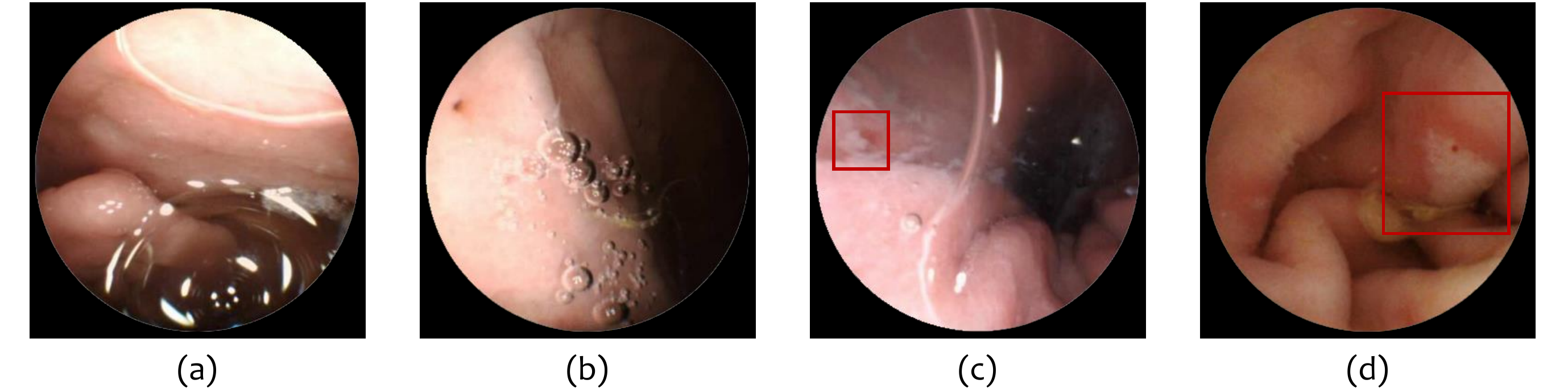}
\end{center}
\caption{
    Four images captured by WCE. (a) and (b) do not contain gastric ulcers, while (c) and (d) contain ulcers of different sizes (marked by a red rectangle).
}
\label{Fig:Sample}
\end{figure}

In this paper we consider gastric ulcer, one of the most common GI diseases that can be seen in WCE images~\cite{Eid2013}. We collected a large dataset with more than $3$ million images, but within it, only $1\%$ of them were annotated with lesion areas around the ulcer. As shown in Figure~\ref{Fig:Sample}, the size of the lesion area may vary, but most lesions are very small. There are, in general, two ways of learning from these data, {\em i.e.}, training a classifier on {\em local} patches and {\em global} images, respectively. Both models have their advantages and disadvantages. Local models are sensitive to small lesions, but can also produce a lot of false alarms. Global models run much faster, but suffer from a relatively low classification accuracy.

To take both benefits and avoid drawbacks, we present the {\bf infinite curriculum learning} approach. The core idea is the same as curriculum learning~\cite{Bengio2009}, {\em i.e.}, ranking the training samples according to their difficulties, training a classifier first on simple patches, and then adapting it gradually to the entire images. Our key innovation is to generalize this approach to an {\bf infinite} sampling space by a reasonable approximation, which {\em only} uses the {\em scale} of each patch to measure its difficulty. Our approach achieves a binary classification accuracy of $87\%$ (part of false-positives were caused by mis-annotation) on a large-scale dataset, which significantly outperforms the baseline. Applied to a real-world scenario, our approach reduces the workload of a physician by $90\%$--$98\%$.

\section{Our Approach}
\label{Approach}

\subsection{Background and Motivation}
\label{Approach:Motivation}

Let our training data be a finite set ${\mathcal{S}}={\left\{\left(\mathbf{X}_n,y_n,\mathcal{R}_n\right)\right\}_{n=1}^N}$ where $N$ is the number of samples. For each $n$, $\mathbf{X}_n$ is a $480\times480\times3$ matrix (where the last dimension indicates the RGB channels) and ${y_n}\in{\left\{0,1\right\}}$ is the label, {\em i.e.}, whether gastric ulcers can be observed from $\mathbf{X}_n$. If ${y_n}={1}$, {\em i.e.}, gastric ulcers are detected, then $\mathcal{R}_n$ contains several bounding boxes, each of which covers a lesion area, otherwise ${\mathcal{R}_n}={\varnothing}$. The goal is to design a function ${\tilde{y}}={f\!\left(\mathbf{X};\boldsymbol{\theta}\right)}$ to classify each testing image $\mathbf{X}$, where $\boldsymbol{\theta}$ is the model parameters. In the context of deep learning, $f\!\left(\cdot;\boldsymbol{\theta}\right)$ is a hierarchical neural network, and $\boldsymbol{\theta}$ is the weights, mostly related to the convolutional layers.

More specifically, we train a $101$-layer deep residual network~\cite{he2016deep} as $f\!\left(\cdot;\boldsymbol{\theta}\right)$. This network takes an $224\times224$ input image, uses one $7\times7$ convolutional layer and $33$ channel bottleneck modules ($99$ layers in total) to down-sample the image to $7\times7$, and finally applies one fully-connected layer for classification. There is a residual connection between the input and output of each bottleneck, which accelerates convergence and enables us to train it from scratch.

However, there are still two major challenges. First, due to the way of data annotation (detailed in Section~\ref{Experiments:Dataset}), the number of positive training samples is very limited. Second, when the capsule is moving within the stomach, the camera may capture the gastric ulcer in various viewpoints, resulting a large variation in the scale and position of the lesion area. Most gastric ulcers annotated by the physicians are of a small size, {\em i.e.}, $85\%$ of the annotated lesions are smaller than $1/25$ of the entire image ({\em e.g.}, Figure~\ref{Fig:Sample} (c)). However, detecting a large lesion ({\em e.g.}, Figure~\ref{Fig:Sample} (d)) is also important in diagnosis. These difficulties motivate us to design an approach which first learns from small patches, and then gradually adjusts itself to the entire image\footnote{There is another choices, {\em i.e.}, training a detection model to localize the lesion areas. But, a detection network runs much ($5\times$) slower at the testing stage, bringing heavy computational costs especially when a patient has roughly $3\rm{,}000$ images to test. In addition, the limited number of positive images as well as gastric ulcer boxes largely downgrades the performance of a detection network.}. We use the curriculum learning framework~\cite{Bengio2009} to implement this idea.

\subsection{Curriculum Learning in an Infinite Sampling Space}
\label{Approach:CurriculumLearning}

We start with sampling patches. In each case $\left(\mathbf{X}_n,y_n,\mathcal{R}_n\right)$, we randomly sample a disc with a diameter in $\left[96,480\right]$ (the diameter of $\mathbf{X}_n$ is $480$) which covers all lesion areas in $\mathcal{R}_n$, if ${\mathcal{R}_n}\neq{\varnothing}$. The entire image $\mathbf{X}$ is always sampled as a patch -- this is to confirm that we preserve all full-image training data, as the final goal is full-image classification. We denote each sampled patch by $\left(\mathbf{X}_m',y_m'\right)$. Note that the sampling space is an {\bf infinite} set.

Then, we define a function $c\!\left(\mathbf{X}_m',y_m',\boldsymbol{\theta}\right)$ to compute the {\em complexity} of each training sample $\left(\mathbf{X}_m',y_m'\right)$. It is a weighted sum of two terms, {\em i.e.}, ${c\!\left(\mathbf{X}_m',y_m',\boldsymbol{\theta}\right)}={r\!\left(\mathbf{X}_m'\right)+\lambda\cdot\left|y_m'-f\!\left(\mathbf{X}_m';\boldsymbol{\theta}\right)\right|^2}$, where $r\!\left(\mathbf{X}_m'\right)$ and $\left|y_m'-f\!\left(\mathbf{X}_m';\boldsymbol{\theta}\right)\right|^2$ are named the {\em image complexity} and {\em classification complexity} of $\left(\mathbf{X}_m',y_m'\right)$, respectively. $r\!\left(\mathbf{X}_m'\right)$ is larger when $\mathbf{X}_m'$ is more complex. Then, we partition the training set into $K$ subgroups ${\mathcal{S}'}={\mathcal{S}_1'+\mathcal{S}_2'+\ldots+\mathcal{S}_K'}$, and guarantee that for ${1}\leqslant{k_1}<{k_2}\leqslant{K}$, we have ${c\!\left(\mathbf{X}_{m_1}',y_{m_1}',\boldsymbol{\theta}\right)}<{c\!\left(\mathbf{X}_{m_2}',y_{m_2}',\boldsymbol{\theta}\right)}$, $\forall{\left(\mathbf{X}_{m_1}',y_{m_1}'\right)}\in{\mathcal{S}_{k_1}'}$ and $\forall{\left(\mathbf{X}_{m_2}',y_{m_2}'\right)}\in{\mathcal{S}_{k_2}'}$. The training process is composed of $K$ {\em stages}, and the model is allowed to see the data in $\bigcup_{l\leqslant k}\mathcal{S}_l'$ at the $k$-th stage, {\em i.e.}, the complexity of training data goes up gradually.

To better fit our problem, we make two modifications. First, we define the {\em image complexity} $r\!\left(\mathbf{X}_m'\right)$ to be proportional to the scale of $\mathbf{X}_m'$, {\em e.g.}, the diameter of $\mathbf{X}_m'$ divided by $480$ (the maximal diameter). This aligns with our motivation, {\em i.e.}, training the model on small (and so easy) patches first, and then tuning it on large (and so difficult) cases. The {\em classification complexity} $\left|y_m'-f\!\left(\mathbf{X}_m';\boldsymbol{\theta}\right)\right|^2$ depends on the parameters $\boldsymbol{\theta}$, and thus needs to be updated with training. But, note that there are an infinite number of patches, so computing this function on each of them becomes intractable. An approximate and efficient solution is to switch off this term, allowing the complexity of a patch to be completely determined by its size. In fact, we find that for any two patches $\left(\mathbf{X}_{m_1}',y_{m_1}'\right)$ and $\left(\mathbf{X}_{m_2}',y_{m_2}'\right)$, the probability of ${\left|y_{m_1}'-f\!\left(\mathbf{X}_{m_1}';\boldsymbol{\theta}\right)\right|^2}<{\left|y_{m_2}'-f\!\left(\mathbf{X}_{m_2}';\boldsymbol{\theta}\right)\right|^2}$ is close to $1$ when $r\!\left(\mathbf{X}_{m_2}'\right)-r\!\left(\mathbf{X}_{m_1}'\right)$ is sufficiently large, {\em e.g.}, greater than $1/5$. Therefore, we partition the sampled patch set $\mathcal{S}'$ into ${K}={5}$ subsets. The first one, $\mathcal{S}_1$, contains all patches satisfying ${r\!\left(\mathbf{X}_m'\right)}\in{\left[1/5,2/5\right)}$. Then $\mathcal{S}_2$, $\mathcal{S}_3$ and $\mathcal{S}_4$ contain those within $\left[2/5,3/5\right)$, $\left[3/5,4/5\right)$ and $\left[4/5,1\right)$, respectively, and the final one, $\mathcal{S}_5$, contains all full images, {\em i.e.}, ${r\!\left(\mathbf{X}_m'\right)}={1}$. This strategy is named {\bf infinite curriculum learning}.

Second, as the final goal is full-image classification, we do not continue training on the small patches used previously when we enter a new stage. In practice, this strategy improves the stability as well as overall performance of our approach, because the sampling space does not increase with time.

\renewcommand{\figurewidth}{11.0cm}
\begin{figure}[t]
\begin{center}
    \includegraphics[width=\figurewidth]{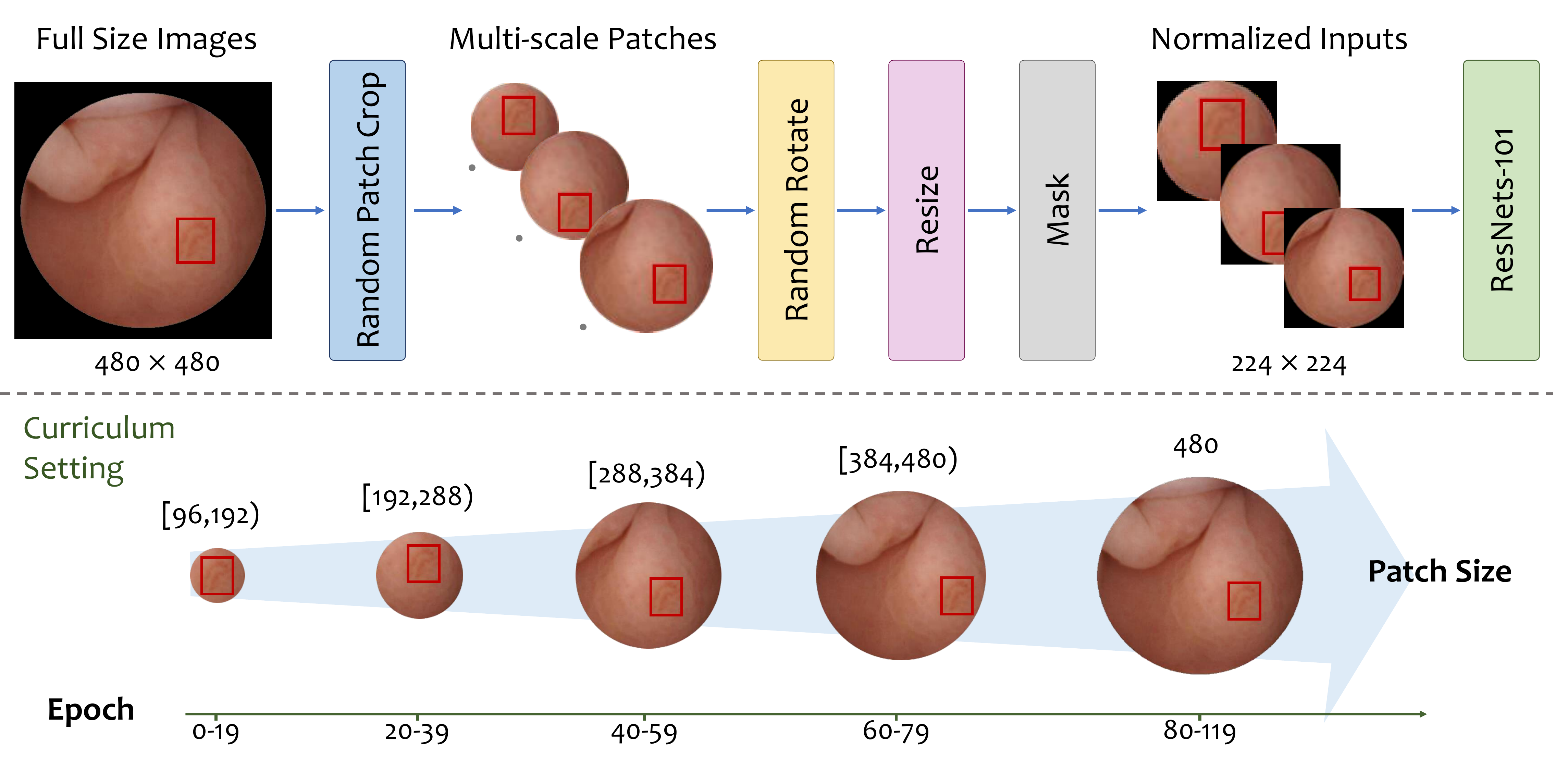}
\end{center}
\caption{
    The curriculum learning process (best viewed in color). All sampled patches contain the lesion area(s). We partition these patches into $5$ subgroups according to their sizes, and use them orderly throughout the training process. Data augmentation is applied to each patch before it is fed into the deep network.
}
\label{Fig:framework}
\end{figure}

The training process is illustrated in Figure~\ref{Fig:framework}. We perform a total of $120$ epochs. Each epoch contains $\sim170$ iterations with a mini-batch size of $256$. The base learning rate is $0.1$, and divided by a factor of $10$ after $40$ and $80$ epochs. The five subsets $\mathcal{S}_1$ through $\mathcal{S}_5$ are used orderly. Except for $\mathcal{S}_5$ which is used in the last $40$ epochs, all others are used in $20$ epochs. We train on the full images for a longer stage to improve the stability in the testing stage. Note that all subsets are infinite, so we sample patches in an online manner. Each training image is randomly rotated by ${\alpha}\in{\left[0^\circ,360^\circ\right)}$ (data augmentation), resized into $224\times224$, masked (all pixels within the minimal square that covers it but outside the disc are set to be black), and finally fed into the deep network.

\section{Experiments}

\subsection{Dataset and Evaluation}
\label{Experiments:Dataset}

\newcommand{\colwidthA}{1.35cm}
\begin{table}[t]
\centering
\begin{tabular}{|l||R{\colwidthA}|R{\colwidthA}|R{\colwidthA}|R{\colwidthA}||R{\colwidthA}|}
\hline
{}                             &              Fold \#0 &              Fold \#1 &              Fold \#2 &              Fold \#3 &                 Total \\
\hline\hline
Ulcer patients                 &                 $382$ &                 $382$ &                 $370$ &                 $365$ &          $1\rm{,}499$ \\
\hline
Non-ulcer patients             &                  $78$ &                  $78$ &                  $90$ &                  $93$ &                 $339$ \\
\hline
Total patients                 &                 $460$ &                 $460$ &                 $460$ &                 $458$ &          $1\rm{,}838$ \\
\hline\hline
Images from ulcer patients     &        $778\rm{,}638$ &        $763\rm{,}364$ &        $827\rm{,}597$ &        $755\rm{,}107$ & $3\rm{,}124\rm{,}706$ \\
\hline
Images from non-ulcer patients &        $145\rm{,}911$ &        $156\rm{,}755$ &        $213\rm{,}552$ &        $250\rm{,}011$ &        $766\rm{,}229$ \\
\hline
Images with gastric ulcers     &          $8\rm{,}523$ &          $8\rm{,}991$ &          $8\rm{,}366$ &          $8\rm{,}063$ &         $33\rm{,}943$ \\
\hline
Total Images                   &        $924\rm{,}549$ &        $920\rm{,}119$ & $1\rm{,}041\rm{,}149$ & $1\rm{,}005\rm{,}118$ & $3\rm{,}890\rm{,}935$ \\
\hline
\end{tabular}
\caption{
    The numbers of patients and images in each fold. The images annotated with gastric ulcers only occupy a very small fraction of those from ulcer patients.
}
\label{Tab:Dataset}
\end{table}

We collected a dataset of Wireless Capsule Endoscopy (WCE) images for gastric ulcer classification. As far as we know, there is no public dataset for this purpose. Detailed statistics are summarized in Table~\ref{Tab:Dataset}. Our dataset is collected from $1\rm{,}838$ patients. Among them, $1\rm{,}499$ have at least one frame annotated with gastric ulcers. On all ulcer frames, the physicians manually labeled a bounding box for each lesion area. The physicians annotated with confidence, {\em i.e.}, only those regions containing enough information to make decisions were annotated, which implies that part of ulcer frames were not annotated (see Figure~\ref{Fig:Cases} for examples).

The set of positive samples contains the $33\rm{,}943$ images labeled with gastric ulcers, and the negative set contains all $766\rm{,}229$ images from the $339$ non-ulcer patients. We randomly split the patients into $4$ folds of approximately the same size, and perform standard cross-validation. In all testing sets, there are roughly the same number of positive and negative cases. We report the average classification accuracy as well as the area under the ROC curve (AUC).

\subsection{Quantitative Results}
\label{Experiments:Results}

We denote our approach by {\tt CURRIC\_5}, {\em i.e.}, the $5$-stage infinite curriculum learning approach described in Section~\ref{Approach:CurriculumLearning}. To verify its effectiveness, we consider four other learning strategies with the same number of epochs, learning rates, {\em etc}. The first one is named {\tt RANDOM}, which samples a patch of a random size in $\left[96,480\right]$ throughout the entire training process. The second one, named {\tt FULL}, instead fixes the patch size to be $480$. These two variants simply use one training stage without taking the benefits of curriculum learning. The third and fourth variants reduce the number of stages by combining the first four stages and the last two stages, leading to a $2$-stage process ({\tt CURRIC\_2}) and a $4$-stage process ({\tt CURRIC\_4}), respectively. By combining several stages, we keep the total number of epochs, and use the union of the original subsets in the combined stage.

\renewcommand{\colwidthA}{0.9cm}
\newcommand{\colwidthB}{0.9cm}
\begin{table}[t]
\centering
\begin{tabular}{|l||R{\colwidthA}|R{\colwidthA}|R{\colwidthA}|R{\colwidthA}|R{\colwidthA}||R{\colwidthA}|R{\colwidthA}|R{\colwidthA}|R{\colwidthA}|R{\colwidthA}|}
\hline
\multirow{2}{*}{Approach} & \multicolumn{5}{c||}{Classification Accuracy}
                          & \multicolumn{5}{c|}{AUC Value}                                                               \\
\cline{2-11}
{}                        &            F \#0 &            F \#1 &            F \#2 &            F \#3 &              AVG
                          &            F \#0 &            F \#1 &            F \#2 &            F \#3 &              AVG \\
\hline\hline
{\tt RANDOM}              &          $83.79$ &          $83.65$ &          $82.32$ &          $82.51$ &          $83.07$
                          &          $92.46$ &          $92.21$ &          $92.38$ &          $90.35$ &          $91.85$ \\
\hline
{\tt FULL}                &          $85.51$ &          $84.23$ &          $86.74$ &          $83.55$ &          $85.01$
                          &          $93.36$ &          $91.28$ &          $93.62$ &          $90.87$ &          $92.28$ \\
\hline
{\tt CURRIC\_2}           & $\mathbf{87.26}$ &          $85.57$ &          $86.73$ &          $85.73$ &          $86.32$
                          & $\mathbf{94.12}$ & $\mathbf{93.07}$ &          $93.82$ &          $92.57$ &          $93.40$ \\
\hline
{\tt CURRIC\_4}           &          $85.78$ &          $85.45$ &          $88.66$ &          $85.71$ &          $86.40$
                          &          $93.30$ &          $92.09$ &          $\mathbf{95.34}$ &          $93.05$ &          $93.44$ \\
\hline
{\tt CURRIC\_5}           &          $87.05$ & $\mathbf{86.07}$ & $\mathbf{88.83}$ & $\mathbf{86.12}$ & $\mathbf{87.02}$
                          &          $94.02$ &          $92.81$ & $95.33$ & $\mathbf{93.17}$ & $\mathbf{93.83}$ \\
\hline
\end{tabular}
\caption{
    Classification accuracy ($\%$) and area-under-curve (AUC, $\%$) values for each fold and with respect to different learning strategies.
}
\label{Tab:Results}
\end{table}

\renewcommand{\figurewidth}{12cm}
\begin{figure}[t]
\begin{center}
    \includegraphics[width=\figurewidth]{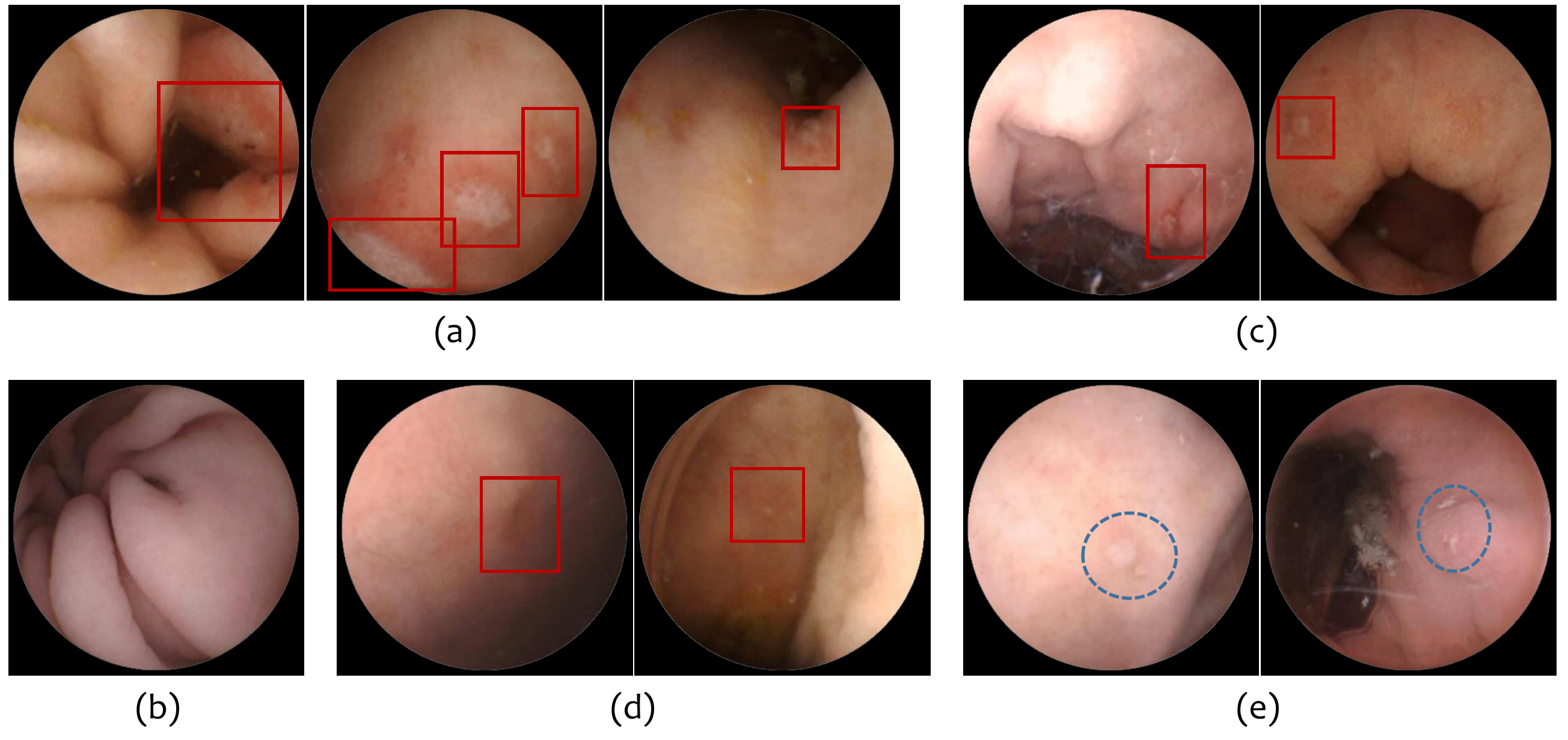}
\end{center}
\caption{
    Examples of gastric ulcer classification. (a) and (b) are true positive and true negative cases. (c) shows two ``false positives'' which are actually mis-annotated by the physicians. (d) and (e) are false negative and false positive cases. Solid rectangles and dashed circles mark real gastric ulcers and false alarms, respectively.
}
\label{Fig:Cases}
\end{figure}

Results are summarized in Table~\ref{Tab:Results}. We can first conclude that the {\tt CURRIC\_5} approach achieves the highest overall performance, {\em i.e.}, a $87.02\%$ classification accuracy and a $93.83\%$ AUC value. Both these numbers are significantly higher than {\tt FULL} ({\em e.g.}, $13.41\%$ relative classification error drop), the direct baseline, indicating the effectiveness of curriculum learning. By contrast, the {\tt RANDOM} approach produces much lower (yet unstable among different epochs) accuracies using completely random sampling. This is partly because the training process does not align with the final goal, which is full image classification. This also explains the improvement from {\tt CURRIC\_4} to {\tt CURRIC\_5}. When we use the last $40$ epochs to adjust the {\tt RANDOM} model ({\em i.e.}, {\tt CURRIC\_2}), much higher performance is achieved. In addition, the advantage of {\tt CURRIC\_2} over {\tt FULL} also implies the importance of using random-sized patches to capture multi-scale information, and that of {\tt CURRIC\_5} over {\tt CURRIC\_2} suggests that curriculum learning is an effective way of controlling random data distributions in the training process.

Some classification examples are shown in Figure~\ref{Fig:Cases}. Our approach can detect some positive samples which were not annotated by the physicians.

\subsection{Application to Gastric Ulcer Screening}
\label{Experiments:Application}

\renewcommand{\figurewidth}{12cm}
\begin{figure}[t]
\begin{center}
    \includegraphics[width=\figurewidth]{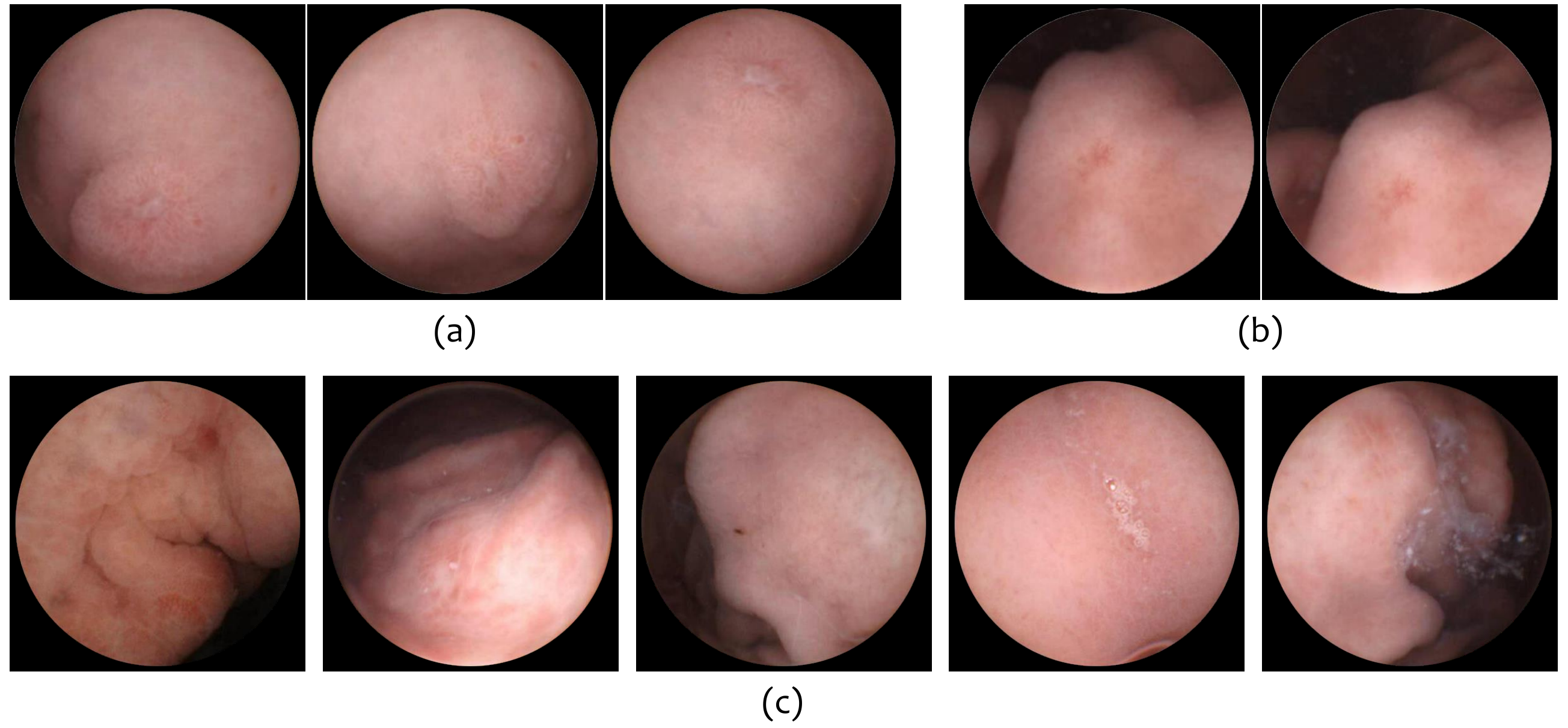}
\end{center}
\vspace{-0.5cm}
\caption{
    Applying our approach for gastric ulcer screening. The top and bottom rows display frames from two patients, the top one suffers from gastric ulcer but the bottom one does not. (a) and (b) show some high-confidence (${y}\geqslant{0.8}$) and medium-confidence (${0.6}\leqslant{y}<{0.8}$) examples, respectively, and all these images were verified to be true-positives. (c) contains the top-ranked images predicted by our approach. The physician verified that all these cases are false-positives, but these images are worth investigation.
}
\label{Fig:Scenario}
\end{figure}

To apply our approach to gastric ulcer screening, we first notice several facts. First, the major problem that bothers the physicians is the large number of frames to be checked. Second, a gastric ulcer patient often has a number of positive frames, and detecting any one of them is able to diagnose the presence of gastric ulcers. Therefore, we modify the classification threshold, setting $0.8$ to be a high threshold, and $0.6$ to be a medium threshold\footnote{In comparison, in the classification task, all the thresholds were set to be $0.5$.}. The physician first checks all high-confidence frames; if no ulcers are detected, he/she turns to check the medium-confidence ones. If still nothing suspicious is found, this patient is considered free of gastric ulcers. Using this system, we reduce the number of frames to check for each patient from roughly $3\rm{,}000$ to $60$--$300$, which largely saves time for a physician to screen gastric ulcers. The average time for our system to process $3\rm{,}000$ frames does not exceed $40$ seconds.

\section{Conclusions}
\label{Conclusions}

This paper presents an approach, named infinite curriculum learning, for classifying WCE images according to the presence of gastric ulcers. This is motivated by the requirement of detecting multi-scale, especially small lesions, meanwhile being transferable to full images. To sample from an infinite space, we design an efficient way to compute the complexity of each sample only by its size, leading to a training schedule in which the patch size is gradually increased until reaching the maximum (full image). While being simple, our approach achieves consistent accuracy gain over the baseline. Setting a high classification threshold enables our approach to largely reduce the workload of the physicians.

\bibliographystyle{splncs03}
\bibliography{typeinst}
\end{document}